\documentclass[sn-mathphys]{sn-jnl}
\jyear{2021}%

\theoremstyle{thmstyleone}%
\newtheorem{theorem}{Theorem}
\newtheorem{proposition}[theorem]{Proposition}%
\theoremstyle{thmstyletwo}%
\theoremstyle{thmstylethree}%
\newtheorem{definition}{Definition}%

\usepackage{graphicx}
\usepackage{caption}
\begin{document}

\title[Article Title]{Adaptive perturbation adversarial training: based on reinforcement learning}

\author[1]{\fnm{Zhishen} \sur{Nie} }\email{12019202386@mail.ynu.edu.cn} 

\author*[1,2]{\fnm{Ying} \sur{Lin}}\email{Linying@ynu.edu.cn}

\author[1]{\fnm{Sp} \sur{Ren}}

\author[1]{\fnm{Lan} \sur{Zhang}}

\affil*[1]{\orgdiv{ School of Software}, \orgname{ Yunnan University}, \orgaddress{ \city{ Kunming}, \postcode{650500}, \state{Yunnan Province}, \country{China}}}

\affil[2]{\orgdiv{ Key Laboratory in Software Engineering of Yunnan Province}}


\abstract{Adversarial training has become the  primary method to defend against adversarial samples. However, it is hard to practically apply due to many shortcomings. One of the shortcomings of adversarial training is that it will reduce the recognition accuracy of normal samples. Adaptive perturbation adversarial training is proposed to alleviate this problem. It uses marginal adversarial samples that are close to the decision boundary but does not cross the decision boundary for adversarial training, which improves the accuracy of model recognition  while maintaining the robustness of the model. However, searching for marginal adversarial samples brings additional computational costs. This paper proposes a method for finding marginal adversarial samples based on reinforcement learning, and combines it with the latest fast adversarial training technology, which effectively speeds up training process and reduces training costs.}


\maketitle

\section{Introduction}\label{sec1}

It has been widely known that DNNs (Deep Neural Networks, DNNs) are susceptible to adversarial samples. Using adversarial samples for adversarial training becomes an effective method to defend against adversarial attacks. FGSM (Fast gradient sign method , FGSM) \cite{FGSM} is the first proposed adversarial training method, known for its fast training speed. However, in the literature \cite{PGD}, it has been proved that this method cannot defend the adversarial samples generated by multiple iterations, and therefore proposed a more effective PGD (Project Gradient Descent, PGD) adversarial training method. PGD adversarial training uses smaller perturbation steps to generate stronger adversarial samples for multiple iterations to enhance the robustness of the model. Although  adversarial training can effectively defend against adversarial samples, there are two main shortcomings: (1) The computational cost is high. Compared with normal training, adversarial training usually costs 2-10 times the computational cost; (2)  In the field of computer vision, although the robustness of the model is improved, all adversarial training methods  will destroy the generalization ability of the model to a certain extent These two shortcomings greatly restrict  its applications.

There have been several works \cite{Free-AT, YOPO} researching how to reduce the computational cost of adversarial training. For example, Fast adversarial training \cite{FAT-RS}  and its subsequent improved version \cite{FAT-Align, kim2020understanding}   can reduce the number of back propagation necessary for adversarial training to 2-3 times. In other words, the fast adversarial training technology only spends 2-3 times the computational cost (compared to  regular training) while the model still obtains good robustness and better generalization ability.

Aiming at the problem that adversarial training impairs the generalization ability of the trained model, more effective adaptive perturbation adversarial training methods \cite{MMA, IAAT} were  proposed. In traditional adversarial training process, each sample was added with the same perturbation step. But  in the adaptive perturbation adversarial training process, the perturbation step  added by each sample is different. This improvement greatly enhances  the generalization ability of the model. However, due to the additional computational cost of searching for marginal adversarial samples, the computational cost is still higher than that of fast adversarial training.

Reinforcement learning (RL) is a technology that imitates  the behavior of living creatures seeking advantages and avoiding disadvantages. Unlike supervised learning, reinforcement learning does not have a fixed method, and everything must be done in interaction with the environment. A reinforcement learning system can be thought of as a simulated world. This system includes elements such as environment, agent, reward, policy, state and so on. The most critical elements are rewards and policy. The reward function in reinforcement learning is equivalent to the loss function in supervised learning. The difference is that the reward function in reinforcement learning needs to be maximized. policy is another key factor, which determines what actions the agent will make based on observations. The cornerstone of reinforcement learning is the Markov decision processes \cite{MDP}, which allows the complex environment to be simplified into a state transition equation to simplify the problem.

Deep reinforcement learning combined with DNNs has also made  significant progress in recent years. Representative works include \cite{DQN, DDPG, TD3, TRPO, PPO} etc. But they still have some shortcomings. For example, it is too sensitive to hyperparameter (such as learning rate), which makes training difficult, or the sample efficiency is too low, causing the agent to have a lot of interaction with the environment. The proposed   SAC (Soft Actor Critic, SAC) \cite{SAC} solves the above problems. It uses the Actor-Critic architecture, uses maximum entropy and random strategies to better explore possible optimal paths, thereby improving generalization capabilities. At the same time, it is an off-policy algorithm that can effectively use historical data, so the sample efficiency is also very high. In addition, SAC is also one of the few reinforcement learning algorithms that can be directly deployed in real robots.

The contributions of this article are as follows: \par
(1) We use reinforcement learning to search for marginal adversarial samples, and theoretically prove the rationality of the method. As far as we know, this is the first application of reinforcement learning in searching for marginal adversarial samples. \par
(2) We use the number of calculations as the standard to compare the performance of different methods in calculating marginal adversarial samples in detail. \par
(3) In the past, adaptive perturbation adversarial training was based on PGD, and we adopted the latest fast adversarial training technology, which greatly reduced the computational cost.

\section{Related works}\label{sec2}
\subsection{Adaptive perturbation adversarial training}\label{subsec2}
Adaptive perturbation adversarial training is a recently proposed method. Compared with traditional adversarial training such as PGD adversarial training, its  most prominent feature is to formulate a "unique" perturbation step for each sample.  Its idea is that if a larger perturbation is forcibly added to the samples close to the decision boundary, the model will be forced to change the decision boundary, resulting in poor generalization ability. The MMA is the first proposed adaptive perturbation adversarial training \cite{MMA} and its  optimization goal  is as follows:
\begin{equation}
	 \mathop{min} \limits_{\theta} \{\sum \limits_{i \in S_{\theta}^{+}}  max \{0, d_{max}-d(x_{i}, y_{i}, \theta)\} + \beta \cdot  \sum \limits_{i \in S_{\theta}^{-}}  L(x_{i}, y_{i}, \theta)) \}
\end{equation}
\noindent where $\theta$  is the parameter of the classification neural network, $d_{max}$ is a hyperparameter that represents the maximum value adversarial perturbation, $\beta$ is also a hyperparameter, which is used to balance the two loss terms, $L(\cdot)$  represents the classification loss, usually cross entropy. $S_{\theta}^{+}$ represents the set of samples that are correctly classified, $S_{\theta}^{-}$ represents the set of misclassified samples. MMA maximizes the margins of correctly classified samples and minimizes the classification loss of incorrectly classified samples. In addition, the author also gives an experimental analysis, confirming the effectiveness of adaptive perturbation adversarial training.

\subsection{Fast adversarial training}\label{subsec2}
With the rapid development of adversarial training, one of its drawbacks has gradually emerged: the computational cost of adversarial training is too high, especially the PGD. Therefore, some scholars restarted researchs  of using single-step iterative generation of adversarial samples (such as FGSM) for adversarial training. These methods are collectively referred to as FAT (Fast Adversarial Training, FAT). Free-AT (Free Adversarial Training, Free-AT) \cite{Free-AT} was the first FAT method proposed to reduce computational cost. Compared with PGD, Free-AT uses a larger perturbation step size to generate adversarial samples, and each generated adversarial sample  is used to train the model, thus significantly speeding up the model convergence speed. However,  the same batch of data still needs to be backpropagated multiple times. In the literature [5], the authors proposed a fast adversarial training method. The author uses FGSM + RS (Random Step, RS) instead of PGD to generate adversarial sample. This method fully inherits the advantages of FGSM adversarial training, and does not require multiple backpropagation of the same batch of data, so it significantly saves computational costs.
However, the above-mentioned methods all have such a problem: when the adversarial perturbation is too large, catastrophic overfitting will occur. The model’s accuracy can suddenly drop to zero when confronting PGD attacks, while increase significantly when confronting FGSM attacks. The current solution is to stop training before overfitting, so that the model can obtain sub-optimal robustness.

The better solution to overfitting problem is \cite{FAT-Align}. The authors believed that the key to avoiding catastrophic overfitting is to maintain the consistency of PGD adversarial samples and FGSM adversarial samples, and ensuring consistency lies in the local linearity of the model. They  proposed a regularization of cosine similarity:
\begin{equation}
	\Omega(x,y,\theta) = \mathbb{E}_{(x,y) \thicksim D,\eta  \thicksim \mu([-\varepsilon, \varepsilon])}[1-cos(\nabla_{x}L(x,y,\theta),\nabla_{x}L(x+\eta,y,\theta) ]
\end{equation}\par
It guarantees the local linearity of the model within $\ell_{\infty} \mbox{-}balls$, and thus solves catastrophic overfitting well. The authors also proved  that this method still does not appear catastrophic overfitting even under large perturbations.

\section{Motivation}\label{sec3}
In the previous adaptive perturbation adversarial training, the general process of finding marginal adversarial samples is as follows:\par
$x_t=x_{t-1} + \varepsilon \cdot sign(grad) $ \par
if $f(x_{t})$ is right then $x_{t}=x_{t-1} +\varepsilon \cdot  sign(grad) $ \par
 else $x_t=x_{t-1} - \varepsilon \cdot sign(grad )$ \par
where $grad = \nabla _x L \left( x,y,\theta \right)$.However, this process is a qualitative method rather than a quantitative method. It can only find the sample closest to the decision boundary under the current perturbation step. As for how far the sample is from the decision boundary, it cannot be determined.

The IAAT (Instance Adaptive Adversarial Training, IAAT) \cite{IAAT}  uses fixed perturbation step length. However, the perturbation  step length is  a more difficult value to determine. If a fixed perturbation step size is used, then only a small perturbation step size can be taken, because if the step size is too large, the calculated sample may be far away from the decision boundary. However, the training cost is increased due to too many calculation times for small steps. According to our observations in the experiments, in the iterative process, if the perturbation step length is fixed each time, the closer to the decision boundary, the more dramatic the change in confidence. Confidence is a non-linear relationship in adversarial perturbations and is closely related to many factors, such as data sets and network architectures. Therefore, it is difficult to determine the size of a suitable fixed perturbation step. MMA is based on the binary search to explore the perturbation step size, which saves the calculation cost to a large extent compared with the fixed perturbation step size. The upper bound of the half search is difficult to determine, and an excessively large upper bound will increase the number of search steps. 

From the above analysis, it could be seen that the process of finding marginal adversarial samples is a process of continuous interaction between the model and the sample. Considering reinforcement learning is a process of continuous interaction between the environment and the agent, which  brings us enlightenment: whether we can use reinforcement learning to optimize the calculation process of marginal adversarial samples? 

\section{Problems and solutions}\label{sec4}

The core idea of this paper is to use a larger perturbation when the sample is far from the decision boundary, so that the sample can quickly approach the decision boundary, but  when the sample is close to the decision boundary, a smaller perturbation is used to make it close enough to the decision boundary. In this paper we assumes that the output of all classification neural networks has been normalized by softmax. Using adaptive perturbation setp faces two problems that we must resolve. \\

\textbf{Q1:How to measure the distance  between the sample  and the decision boundary?} \par

Although the absolute distance can be used to measure the distance  between the sample  and the decision boundary:
\begin{equation}
L_{ad}=\underset{\varepsilon}{\min} ( arg\max f( x+\varepsilon \cdot sign(grad) ) \ne k ) 
\end{equation}
where $k$  represents the correct category. But we can’t get the value of  the absolute distance until we calculate it. We introduce the concept of  relative distance, which is easily obtained in the calculation process:
\begin{equation}
L_{rd}=f\left( x+\varepsilon \cdot sign(grad) \right) \left[ k \right] -arg\max  f\left( x+\varepsilon \cdot sign(grad) \right) \left[ i \right] \left( i\ne k \right) 
\end{equation} 
The relative distance can not avoid the influence of some extreme samples, and also can better describe the distance between the sample and the decision boundary. In fact, MMA also used this distance initially, but due to the instability of the first optimization term of formula (1), this distance was not adopted anymore. As we do not directly optimize the neural network with this distance, so this problem can be avoided. \\
\textbf{Q2: How to dynamically adjust the step according to the $L_{rd}$?} \\
The process of calculating marginal samples is nothing more than increasing or reducing perturbation step size. We are keenly aware that the calculation process of marginal adversarial samples based on FGSM satisfies Markov properties.
\begin{proposition}
	For a classification neural network model f with fixed parameters and a sample x,the calculation process of the marginal adversarial sample based on FGSM is a Markov decision process. The formal description is as follows: \par
	\begin{equation}
		f(x_{t}^{adv},y \mid x_{t-1}^{adv}, x_{t-2}^{adv}, ...,x_{0}^{adv}) =f(x_{t}^{adv},y \mid  x_{t-1}^{adv}) 
	\end{equation} 
	
\end{proposition}

The certification process is detailed in Appendix A.1.
We can use any reinforcement learning method to optimize the process of computing marginal adversarial samples. Considering that SAC is stable enough and insensitive to most hyperparameters, we  adopt the architecture of SAC in this article. \par
Algorithm 1 describes the marginal adversarial sample search based on reinforcement learning.\par

\begin{algorithm}
	\caption{Search $\varepsilon$ by Soft Actor-Critic}\label{algo1}
	\begin{algorithmic}[1]
		\Require $f$: the classification neuralnetwork;  $x$: the sample; $y$: the label of $x$; $p$: the policy network of SAC; $\varepsilon_{max}$:  Maximum perturbation step size;	$grad$: the gradient of $x$;$step_{max}$: The maximum number of iterations;
		\State $\varepsilon_{total}, \varepsilon_{last} , step\leftarrow 0$
		\State 	$state_{0} \leftarrow (f(x)[k], f(x)[i], \varepsilon_{total}, \varepsilon_{last})$
	
		\While{$step > step_{max} $ or $ reward > 0$}
		\State $\varepsilon_{last} \leftarrow p(state_{0})$
		\State $\varepsilon_{total} \leftarrow  \varepsilon_{total} + \varepsilon_{last}$
		\State calculate $reward$ by Equation (7)  
		\State $state_{1} \leftarrow (f(x_{adv})[k],f(x_{adv})[i], \varepsilon_{total},  \varepsilon_{last})$
		\State $state_{0} \leftarrow state_{1}$
		\If{the termination conditions are met}
		\State $done = True$
			\State $\varepsilon_{total} \leftarrow clamp(0, \varepsilon_{max}$)
		\Else
		\State $done = False$
		\EndIf
		\State store $state_{0}, state_{1}, reward, done$
		\State $step \leftarrow step + 1$
		\EndWhile
		\State Return  $\varepsilon_{total}$
	\end{algorithmic}
\end{algorithm}
The relevant elements of the reinforcement learning system are as follows: \par
(1) Environment:  We regard Neural network models and samples  as parts of the environment. \par 
(2) State space: the probability label of the sample output by the model (environment) ,the total perturbation step size and the last perturbation step size.
\begin{equation}
	\begin{aligned}
	state = (f(x_{adv})[k], arg \,\, max f(x_{adv})[i],\varepsilon_{total}, \varepsilon_{last} ) \\
	\end{aligned}
\end{equation}
where $ x^{adv} = x + \varepsilon_{total} \cdot sign(grad)$ and $i\neq k $\\
(3) Reward function:
\begin{equation}
reward=
\begin{cases}
	-\frac{1}{1+ e^{-\lvert L_{rd} \rvert} }, & L_{rd}> upper \,\, bound\\
	P (a \, \,  positive \,\, constant), &   low \,\, bound <=L_{rd} <= upper \,\, bound\\
	-\frac{1}{1+ e^{-\lvert L_{rd} \rvert}}, & L_{rd} < low \,\, bound\\
\end{cases}
\end{equation} \par
There is $ lower \,\, bound \leqslant 0 \leqslant upper \,\, bound$. The working mechanism is as follows: before finding the perturbation step that meets the requirements, each step will receive a negative reward, until the step that meets the conditions is found or the maximum number of iterations is exceeded.

The design of the reward function follows two principles: \\
\textbf{Avoid sparse rewards.}
Reward sparseness refers to the fact that it is difficult for the agent to receive a clear reward in most cases during training. The reward function we designed will receive a positive or negative reward at every iteration to avoid sparse rewards. \\
\textbf{Increase the distinction between different positive rewards.} Under the condition of finding the adversarial perturbation satisfying the condition, the computational step size of finding that should be reduced as much as possible.
\begin{definition}[Distinction]
	For any positive real number $a, b, c, d \in \mathbb{R^+}$, If $\frac{a}{b} > \frac{c}{d}$  then it is said that the distinction between $a$ and $b$ is greater than that of $c$ and $d$.
\end{definition}

The formula for calculating the lower bound of $P$ is as follows:

\begin{equation}
	P=\int_{ \lvert lower \,\, bound \lvert}^{1} \frac{1}{1+e^{-L_{rd}}} dL_{rd} + \int_{upper \,\, bound}^{1} \frac{1}{1+e^{-L_{rd}}} dL_{rd}
\end{equation} 
It can be ensured that there will not be a situation where a perturbation that satisfies the condition is found but the total reward is negative.
\begin{proposition}
	Taking into account the degree of discrimination, formula (8) is the optimal value of $P$.
\end{proposition} 

\noindent (4) Action space: Here we only need to add or subtract the steps to complete the one-dimensional action space. What we need to do is to limit the upper and lower limits of the action, that is, the upper and lower limits of adversarial perturbation step size.  \par 

The above are some settings of the key elements of reinforcement learning. In the actual training process, because the parameters of the model are constantly updated, the policy network needs to be trained when it cannot work well. But according to our observations in the experiment, the model needs to be updated hundreds of times, and the policy network only needs to be updated a few times. \par 
The final loss function is as follows:
\begin{equation}
\mathcal{L} \left( x,y,\theta \right) =L \left( x_{adv},y,\theta \right) +\lambda \cdot \varOmega \left( x,y,\theta \right)
\end{equation} 
The reason why  we did not use the loss functions used in MMA and IAAT is that we sure that the generated marginal adversarial samples are close enough to the decision boundary.  The network based on regular cross-entropy loss could learn enough information from the samples. In fact, our experiments have also verified this hypothesis. \par
The pseudo code is shown in Algorithm 2.

\begin{algorithm}
	\caption{Adaptive perturbation adversarial training}\label{algo2}
	\begin{algorithmic}[1]
		\Require $f$: the classification neural network; $X$: the train set; $p$: The policy network of SAC; $Epoch$: The maximum number of iterations; 
	
		\For{ $i = 1$; $i<Epoch$; $i++$}
		\State Read minibatch data $B_{(x,y)} \subset X$ 
		\State $grad \leftarrow \nabla_{x}l(x,y,f)$
		\For{$(x_{i}, \ y_{i}) \ in \ B_{(x,y)} $}
		\State Calculate $\varepsilon_{i} $ by Algorithm 1 
		\EndFor
		\State $x_{adv} \leftarrow x +  \varepsilon \cdot sign(grad)$
		\State $rate_{hit} \leftarrow $  calculate by Equation(10)
		\If{$rate_{hit} < rate_{min} $}
		\State retrain policy networks
		\EndIf
		\State update $f$ by  Equation(9)
		\EndFor
	\end{algorithmic}
\end{algorithm}

\section{Experiments Design}\label{sec5}

\subsection{Fixed parameter classification model}

We first conduct an experiment on a fixed-parameter model, that is , the parameters of the classification model will not be updated. and  only the model in sac will be updated. The data set used here is MNIST \cite{MNIST}.\par 
\textbf{Measurement matric.} We use the hit rate to measure the pros and cons of the method.If the relative distance of a marginal adversarial sample satisfies $0 \leqslant L_{rd}  \leqslant 0.1$ , we call it a hit. The formula for calculating the hit rate is as follows:
\begin{equation}
hit\,\,rate=\frac{x_{success}}{x_{total}}
\end{equation} 
Where  $x_{success}$ represents the hit sample, $x_{total}$ is the total number of samples calculated. We only perform calculations on samples correctly identified by the model. Considering that it is possible to get the same reward multiple times in the early training stage, the value of P in actual training is set as follows:
\begin{equation}
	P_{\min}=step_{\max}\cdot \frac{1}{1 + e^{-1}} 
\end{equation} 
Figure 1 shows the change in the hit rate of policy networks during the training process. 
The structure of the classification network for MNIST is shown in Table 6. What needs to be pointed out here is that the abscissa is the number of batch size instead of the number of epochs. We compared the impact of different $P$ values on the training process. The maximum number of iterations for each sample is 10 times. Generally speaking, a smaller $P$ will increase the upper limit of performance after convergence, which is consistent with our theoretical derivation. However, from the experimental results, an appropriate increase of this value will help speed up the training, but the impact of an  extreme $P$ value on training is disastrous.

\begin{figure}
    \centering
	\includegraphics[width=1\textwidth]{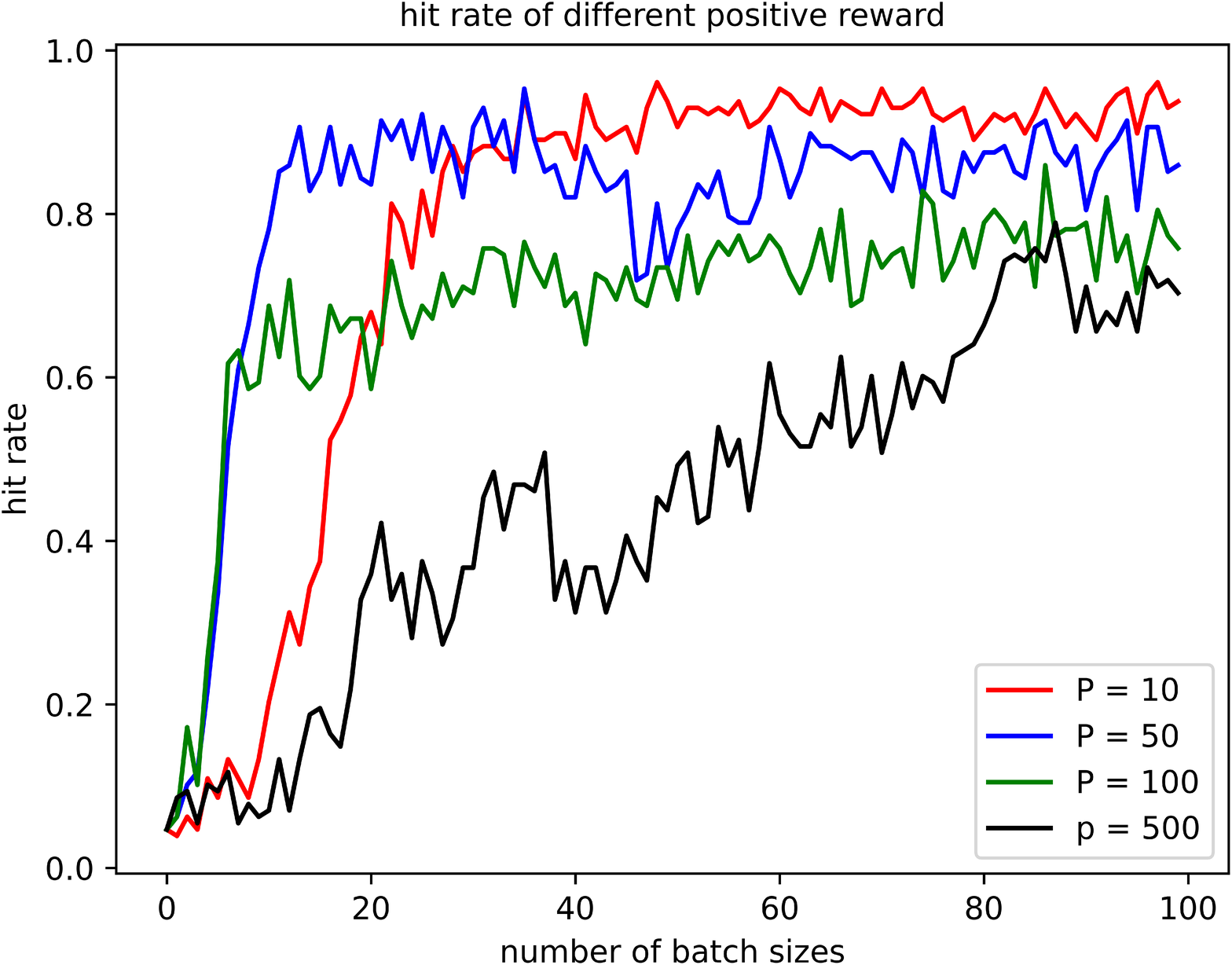}
	\caption{ Hit rate of different positive reward.}
\end{figure}

We also compared different methods, as shown in Table 1. The data in Table 1 is calculated using the MNIST test set data, which has a total of 10,000 samples. Fixed policy parameters during testing. Set p=10. And the policy networks trained with train data of MNIST for 1 epoch. The number of correctly identified samples is 9914. The SAC hyperparameter settings of all experiments are shown in Table 5. Using this parameter in all experiments.

\begin{table}
	\caption{Comparison of different methods.}\label{tab1}
	\centering
	\resizebox{\textwidth}{!}{
		\begin{tabular}{ccccc}
			\toprule
			method & Number of hits   &  Total number of calculations  & Hit rate\\
			\midrule
			     $\varepsilon_{step}=0.03$ ($step_{max}$ = 10) & 708 & 95890 & 7.14\%  \\
				 $\varepsilon_{step}=0.01$ ($step_{max}$ = 30) & 2015 & 271160 & 20.32\%  \\
				 Binary search($step_{max}$ = 5)  & 4110 & 45567 & 41.36\%  \\
				 Binary search ($step_{max}$ = 10)  & 8444 & 59176 & 85.17\%  \\
				 RL (ours) ($step_{max}$ = 5) & 8765 & 32337 & 88.41\%  \\
				RL (ours) ($step_{max}$ = 10) & 9599 & 35398 & 96.82\%  \\
			\bottomrule
		\end{tabular}
	}
\end{table}

\subsection{Adversarial training}
We tested on two data sets, namely Cifar10 \cite{CIFAR-10} and SVHN \cite{SVHN}, and the used classification model  is PreAct-ResNet-18 \cite{Pre-Act_Resnet-18}. The parameter settings are shown in Table 4. Classification neural network uses SGD optimizer, momentum=0.9 and weight decay=5$\times 10^{-4}$. The adjustment method of learning rate is cyclical learning rates \cite{CLR}. The Cifar10 data set adjusts the learning rate within 1-30 epochs, SVHN adjusts the learning rate within 1-15 epochs, and trains at a learning rate of 1e-3 in the remaining epochs. If the average hit rate of the last 10 batches is less than 90\%, the policy network will be trained. 

The experimental results on the cifar10 data set are shown in Table 2. Compared with MMA, our method has similar performance, especially when PGD attack iterations are less than 20.  Our proposed method  only requires 40 epochs to reach convergence, however, the MMA requires 50,000 epochs to reach convergence. Therefore, considering the number of back propagation and the epochs required for training, the training cost of our proposed  method is lower than that of MMA.

\begin{table}
	\caption{Experimental results of different adversarial training on Cifar10.}\label{tab2}
	\centering
	\resizebox{\textwidth}{!}{
		\begin{tabular}{ccccc}
			\toprule
			\multirow{2}{*}{Model}	  &  \multicolumn{4}{c}{Attack}   \\
			                          & None & Pgd-10 & Pgd-20 & Pgd-50 \\
			\midrule
		    Standard   & 94.03\% & 0.00\% & 0.00\% & 0.00\% \\
		    FGSM +align ($\varepsilon$ = 8/255)   & 81.80\% & 50.83\% & 49.04\% & 48.14\%  \\
		    Free AT ($\varepsilon$ = 8/255)   & 84.09\% & 49.06\% & 49.04\% & 47.12\% \\
		    PGD-10 AT ($\varepsilon$ = 8/255)   & 80.86\% & 52.05\% & 51.16\% & 50.74\% \\
		    MMA ($\varepsilon_{max}$=12/255)  & 88.59\% & 49.05\% &  46.32\%& 45.10\%  \\
		    MMA ($\varepsilon_{max}$=20/255)  & 84.36\% & 54.94\% &  52.87\% & 51.69\%  \\
		    Ours ($\varepsilon_{max}$=8/255)  & 87.52\%  & 50.86\% &  46.60\% & 44.03\% \\
		    Ours ($\varepsilon_{max}$=10/255)  & 85.80\%  & 55.73\% &  52.09\%& 49.78\% \\
			\bottomrule
		\end{tabular}
	}
\end{table}

The experimental results on the SVHN data set are shown in Table 3. SVHN is a data set that are difficult to train. If a constant perturbation step size is applied to samples during the training process, then the training will fail. Our proposed method, free-AT and FGSM + align prematurely converge to a local optimal value, and the final recognition rate is  about 23\%. Only PGD-AT  does not be affected. Therefore, in the first five epochs of training, we  gradually increase the perturbation steps size from 0 to 8/255 or 10/255. As for Free-AT, we use normal training in the first five epochs, and then switch to Free-AT training.

\begin{table}
	\caption{Experimental results of different adversarial training on SVHN.}\label{tab3}
	\centering
	\resizebox{\textwidth}{!}{
		\begin{tabular}{ccccc}
			\toprule
			\multirow{2}{*}{Model}	  &  \multicolumn{4}{c}{Attack}   \\
			& None & Pgd-10 & Pgd-20 & Pgd-50 \\
			\midrule
			Standard   & 96.00\% & 1.37\% & 0.67\% & 0.50\% \\
			FGSM +align ($\varepsilon$ = 8/255)   & 91.13\% & 52.93\% & 49.70\% & 47.95\%  \\
			Free AT ($\varepsilon$ = 8/255)   & 82.43\% & 43.16\% & 41.18\% & 40.14\% \\
			PGD-10 AT ($\varepsilon$ = 8/255)   & 91.31\% & 59.77\% & 57.97\% & 57.20\% \\
			Ours ($\varepsilon_{max}$=8/255)  & 94.40\%  & 55.14\% &  48.60\% & 43.37\% \\
			Ours ($\varepsilon_{max}$=10/255)  & 93.96\%  & 56.45\% &  49.65\%& 45.46\% \\
			\bottomrule
		\end{tabular}
	}
\end{table}

\begin{table}
	\caption{Hyperparameter settings.}\label{tab4}
	\centering
	\resizebox{\textwidth}{!}{
		\begin{tabular}{ccccc}
			\toprule
			Dataset	  &   Step size  &  Learning rate  & $\Lambda$ & epoch  \\
			\midrule
			\multirow{2}{*}{Cifar10}  & $\varepsilon_{max}$ = 8/255  & \multirow{2}{*}{  $ lr_{max}=0.3,lr_{min}=1 \times 10^{-3}$ }  & 0.2 & \multirow{2}{*}{40}  \\
			
		  	 &  $\varepsilon_{max}$ = 10/255   &   &  0.356   &   \\
		  	 
		  	 \multirow{2}{*}{SVHN}  & $\varepsilon_{max}$ = 8/255  & \multirow{2}{*}{  $ lr_{max}=0.15,lr_{min}=1 \times 10^{-3}$ }   & 2.5 & \multirow{2}{*}{20}  \\
		  	 
		  	 &  $\varepsilon_{max}$ = 10/255   &   &  2.812   &   \\
			\bottomrule
		\end{tabular}
	}
\end{table}

\begin{table}
	\caption{ Hyperparameter of SAC.}\label{tab5}
	\resizebox{\textwidth}{!}{
		\begin{tabular}{cc}
			\toprule
			Hyperparameter	  &   Value   \\
			\midrule
	        Batch size &  512\\
	        Critic network& (1024,1024) \\
	        Actor (Policy) network &  (1024,1024) \\
	        Learning rate &  $3 \times 10^{-4}$  \\ 
	        Optimizer & Adam \\
	        Replay buffer size  & $1 \times 10^{6}$ \\
	        Exploration policy &  Uniform Distribution,
	        range (-8/255, 16/255)
	        \\
	        Discount factor  & 0.99 \\ 
	        Target update rate & $5 \times 10^{-3}$ \\

			\bottomrule
		\end{tabular}
	}
\end{table}

\begin{table}[h]\tiny
	\caption{ Classification network architecture for MNIST.}\label{tab5}
	\centering
	\resizebox{0.7\textwidth}{!}{
		\begin{tabular}{c}
			\toprule
			Input: 1 × 28 × 28  image \\
			\midrule
		    4 x 4 conv ,16, stride=2, padding=1  \\
		    Relu  \\
		    4 x 4 conv ,32, stride=2, padding=1 \\
		    Relu \\
		    Flatten \\
		    Linear(1538,100) \\
		    Relu \\
		    Linear(100, 10) \\
		    Output: 1 x 10 tensor \\
			\bottomrule
		\end{tabular}
	}
\end{table}

\newpage
\section{Conclusion}\label{sec6}

This paper in-depth analysis the calculation process of the marginal adversarial samples, and also proposes the use of reinforcement learning to calculate the perturbation step length, which further saves the calculation cost, and can quantitatively calculate the marginal adversarial samples with a specified relative distance, and demonstrates its superior performance through exhaustive experiments. However, in terms of broadly searching the adversarial sample space, the effect of PGD is still better than our proposed method and MMA, Free-AT, etc. Therefore, exploring how to better search the adversarial sample space may be the key to improving the performance of these methods. In addition, reinforcement learning may not be the only way to perturbation step and boundary distance modeling. We believe that if a better method is found, the computational cost of searching for marginal adversarial sample will be further reduced.

\begin{appendices}
	
	\section{Proofs}\label{secA1}
	\subsection{Proof of Proposition 1}
	Proof: For the FGSM-based marginal adversarial sample calculation process:\\
	\begin{equation}
		\begin{aligned}
			& x_{0}^{adv}=x+\varepsilon _0\cdot sign(\nabla _x\ell \left( x,y,\theta \right) )\\
			& x_{0}^{adv}=x+\varepsilon _0\cdot sign( \nabla _x\ell \left( x,y,\theta \right) )\\
			& ......\\
			& x_{t}^{adv}=x_{t-1}+\varepsilon _t\cdot sign(\nabla _x\ell \left( x,y,\theta \right)) \nonumber
		\end{aligned}
	\end{equation}
	
	Apparently 
	\begin{equation}
		x_{t}^{adv}=x+\sum_t^{i=0}{\varepsilon _i}\cdot \nabla _x\ell \left( x,y,\theta \right) 
	\end{equation}

	For any sequence of perturbation steps $\varepsilon _a=\left( \varepsilon _0,\varepsilon _1,...,\varepsilon _{a-1},\varepsilon _a \right) $ and $\varepsilon _b=\left( \varepsilon _0,\varepsilon _1,...,\varepsilon _{b-1},\varepsilon _b \right) $, if $\varepsilon _{a}^{'}\subseteq \varepsilon _a$,$\varepsilon _{b}^{'}\subseteq \varepsilon _b$ and $\sum{\varepsilon _{a}^{'}}=\sum{\varepsilon _{b}^{'}}$, from formula (A1) we know that $x_{\varepsilon _{a}^{'}}^{adv}=x_{\varepsilon _{b}^{'}}^{adv}$. That is to say, the adversarial sample is only related to the sum of the elements of the perturbation sequence, and has nothing to do with the size of a single element of the perturbation sequence and the order of the elements. Therefore, for any sequence of perturbation steps $\varepsilon =\left( \varepsilon _0,\varepsilon _1,...,\varepsilon _{k-1},\varepsilon _k \right) $ there must be: \\
	\begin{equation}
		 f(x_{t}^{adv},y \mid x_{t-1}^{adv},x_{t-2}^{adv}, ...,x_{0}^{adv}) = f(x_{t}^{adv},y \mid x_{t-1}^{adv} ) \nonumber
	\end{equation}
	
	\subsection{Proof of Proposition 2}
	Proof. let $P_{1}$, $P_{2}$, satisfy $P_{1} < P_{2}$.  \\
	For any reward sequence $reward_{1} = {r_{1}, r_{2}, ..., r{j}}$ satisfy  $r_{m} <0. m<j$ and $reward_{2} = {r_{1}, r_{2}, ..., r{k}}$ satisfy  $r_{m} <0. m<k$.\\
	Substitute $P_{1}$, $P_{2}$ into  $r_{j}, r_{k}$. \\
	For $P_{1}$,there is $P_1+\sum_{j-1}^{i=1}{r_i}$, $P_1+\sum_{k-1}^{i=1}{r_i}$. \\
	For $P_{2}$,there is $P_2+\sum_{j-1}^{i=1}{r_i}$, $P_2+\sum_{k-1}^{i=1}{r_i}$. \\
	
	$\frac{P_1+\sum_{j-1}^{i=1}{r_i}}{P_1+\sum_{k-1}^{i=1}{r_i}}-\frac{P_2+\sum_{j-1}^{i=1}{r_i}}{P_2+\sum_{k-1}^{i=1}{r_i}}\\=\frac{\left( P_1+\sum_{j-1}^{i=1}{r_i} \right) \left( P_2+\sum_{k-1}^{i=1}{r_i} \right) -\left( P_2+\sum_{j-1}^{i=1}{r_i} \right) \left( P_1+\sum_{k-1}^{i=1}{r_i} \right)}{\left( P_1+\sum_{k-1}^{i=1}{r_i} \right) \left( P_2+\sum_{k-1}^{i=1}{r_i} \right)}\\=\frac{\left[ P_1P_2+\sum_{j-1}^{i=1}{r_i}\cdot P_2+\sum_{k-1}^{i=1}{r_i}\cdot P_1+\sum_{j-1}^{i=1}{r_i}\cdot \sum_{k-1}^{i=1}{r_i} \right] -\left[ P_1P_2+\sum_{j-1}^{i=1}{r_i}\cdot P_1+\sum_{k-1}^{i=1}{r_i}\cdot P_2+\sum_{j-1}^{i=1}{r_i}\cdot \sum_{k-1}^{i=1}{r_i} \right]}{\left( P_1+\sum_{k-1}^{i=1}{r_i} \right) \left( P_2+\sum_{k-1}^{i=1}{r_i} \right)}\\=\frac{\left( \sum_{j-1}^{i=1}{r_i}\cdot P_2+\sum_{k-1}^{i=1}{r_i}\cdot P_1 \right) -\left( \sum_{j-1}^{i=1}{r_i}\cdot P_1+\sum_{k-1}^{i=1}{r_i}\cdot P_2 \right)}{\left( P_1+\sum_{k-1}^{i=1}{r_i} \right) \left( P_2+\sum_{k-1}^{i=1}{r_i} \right)}\\=\frac{\sum_{j-1}^{i=1}{r_i}\left( P_2-P_1 \right) +\sum_{k-1}^{i=1}{r_i}\left( P_1-P_2 \right)}{\left( P_1+\sum_{k-1}^{i=1}{r_i} \right) \left( P_2+\sum_{k-1}^{i=1}{r_i} \right)}\\=\frac{\left( P_1-P_2 \right) \left( \sum_{k-1}^{i=1}{r_i}-\sum_{j-1}^{i=1}{r_i} \right)}{\left( P_1+\sum_{k-1}^{i=1}{r_i} \right) \left( P_2+\sum_{k-1}^{i=1}{r_i} \right)}$
	
	Because $P_1+\sum_{k-1}^{i=1}{r_i}>0$, $P_2+\sum_{j-1}^{i=1}{r_i}>0$, $P_1<P_2$, and $\sum_{k-1}^{i=1}{r_i}<\sum_{j-1}^{i=1}{r_i}$, so there is $\frac{P_1+\sum_{j-1}^{i=1}{r_i}}{P_1+\sum_{k-1}^{i=1}{r_i}}>\frac{P_2+\sum_{j-1}^{i=1}{r_i}}{P_2+\sum_{k-1}^{i=1}{r_i}}$.
	Therefore, under the premise that the sum of the reward queue is positive, the smaller the P value, the greater the distinction.
\end{appendices}

\bibliography{reference}


\begin{thebibliography}{21}
\ifx \bisbn   \undefined \def \bisbn  #1{ISBN #1}\fi
\ifx \binits  \undefined \def \binits#1{#1}\fi
\ifx \bauthor  \undefined \def \bauthor#1{#1}\fi
\ifx \batitle  \undefined \def \batitle#1{#1}\fi
\ifx \bjtitle  \undefined \def \bjtitle#1{#1}\fi
\ifx \bvolume  \undefined \def \bvolume#1{\textbf{#1}}\fi
\ifx \byear  \undefined \def \byear#1{#1}\fi
\ifx \bissue  \undefined \def \bissue#1{#1}\fi
\ifx \bfpage  \undefined \def \bfpage#1{#1}\fi
\ifx \blpage  \undefined \def \blpage #1{#1}\fi
\ifx \burl  \undefined \def \burl#1{\textsf{#1}}\fi
\ifx \doiurl  \undefined \def \doiurl#1{\url{https://doi.org/#1}}\fi
\ifx \betal  \undefined \def \betal{\textit{et al.}}\fi
\ifx \binstitute  \undefined \def \binstitute#1{#1}\fi
\ifx \binstitutionaled  \undefined \def \binstitutionaled#1{#1}\fi
\ifx \bctitle  \undefined \def \bctitle#1{#1}\fi
\ifx \beditor  \undefined \def \beditor#1{#1}\fi
\ifx \bpublisher  \undefined \def \bpublisher#1{#1}\fi
\ifx \bbtitle  \undefined \def \bbtitle#1{#1}\fi
\ifx \bedition  \undefined \def \bedition#1{#1}\fi
\ifx \bseriesno  \undefined \def \bseriesno#1{#1}\fi
\ifx \blocation  \undefined \def \blocation#1{#1}\fi
\ifx \bsertitle  \undefined \def \bsertitle#1{#1}\fi
\ifx \bsnm \undefined \def \bsnm#1{#1}\fi
\ifx \bsuffix \undefined \def \bsuffix#1{#1}\fi
\ifx \bparticle \undefined \def \bparticle#1{#1}\fi
\ifx \barticle \undefined \def \barticle#1{#1}\fi
\bibcommenthead
\ifx \bconfdate \undefined \def \bconfdate #1{#1}\fi
\ifx \botherref \undefined \def \botherref #1{#1}\fi
\ifx \url \undefined \def \url#1{\textsf{#1}}\fi
\ifx \bchapter \undefined \def \bchapter#1{#1}\fi
\ifx \bbook \undefined \def \bbook#1{#1}\fi
\ifx \bcomment \undefined \def \bcomment#1{#1}\fi
\ifx \oauthor \undefined \def \oauthor#1{#1}\fi
\ifx \citeauthoryear \undefined \def \citeauthoryear#1{#1}\fi
\ifx \endbibitem  \undefined \def \endbibitem {}\fi
\ifx \bconflocation  \undefined \def \bconflocation#1{#1}\fi
\ifx \arxivurl  \undefined \def \arxivurl#1{\textsf{#1}}\fi
\csname PreBibitemsHook\endcsname

\bibitem{FGSM}
\begin{botherref}
\oauthor{\bsnm{Goodfellow}, \binits{I.J.}},
\oauthor{\bsnm{Shlens}, \binits{J.}},
\oauthor{\bsnm{Szegedy}, \binits{C.}}:
Explaining and harnessing adversarial examples.
arXiv preprint arXiv:1412.6572
(2014)
\end{botherref}
\endbibitem

\bibitem{PGD}
\begin{botherref}
\oauthor{\bsnm{Madry}, \binits{A.}},
\oauthor{\bsnm{Makelov}, \binits{A.}},
\oauthor{\bsnm{Schmidt}, \binits{L.}},
\oauthor{\bsnm{Tsipras}, \binits{D.}},
\oauthor{\bsnm{Vladu}, \binits{A.}}:
Towards deep learning models resistant to adversarial attacks.
arXiv preprint arXiv:1706.06083
(2017)
\end{botherref}
\endbibitem

\bibitem{Free-AT}
\begin{botherref}
\oauthor{\bsnm{Shafahi}, \binits{A.}},
\oauthor{\bsnm{Najibi}, \binits{M.}},
\oauthor{\bsnm{Ghiasi}, \binits{A.}},
\oauthor{\bsnm{Xu}, \binits{Z.}},
\oauthor{\bsnm{Dickerson}, \binits{J.}},
\oauthor{\bsnm{Studer}, \binits{C.}},
\oauthor{\bsnm{Davis}, \binits{L.S.}},
\oauthor{\bsnm{Taylor}, \binits{G.}},
\oauthor{\bsnm{Goldstein}, \binits{T.}}:
Adversarial training for free!
arXiv preprint arXiv:1904.12843
(2019)
\end{botherref}
\endbibitem

\bibitem{YOPO}
\begin{botherref}
\oauthor{\bsnm{Zhang}, \binits{D.}},
\oauthor{\bsnm{Zhang}, \binits{T.}},
\oauthor{\bsnm{Lu}, \binits{Y.}},
\oauthor{\bsnm{Zhu}, \binits{Z.}},
\oauthor{\bsnm{Dong}, \binits{B.}}:
You only propagate once: Accelerating adversarial training via maximal
  principle.
arXiv preprint arXiv:1905.00877
(2019)
\end{botherref}
\endbibitem

\bibitem{FAT-RS}
\begin{botherref}
\oauthor{\bsnm{Wong}, \binits{E.}},
\oauthor{\bsnm{Rice}, \binits{L.}},
\oauthor{\bsnm{Kolter}, \binits{J.Z.}}:
Fast is better than free: Revisiting adversarial training.
arXiv preprint arXiv:2001.03994
(2020)
\end{botherref}
\endbibitem

\bibitem{FAT-Align}
\begin{botherref}
\oauthor{\bsnm{Andriushchenko}, \binits{M.}},
\oauthor{\bsnm{Flammarion}, \binits{N.}}:
Understanding and improving fast adversarial training.
arXiv preprint arXiv:2007.02617
(2020)
\end{botherref}
\endbibitem

\bibitem{kim2020understanding}
\begin{botherref}
\oauthor{\bsnm{Kim}, \binits{H.}},
\oauthor{\bsnm{Lee}, \binits{W.}},
\oauthor{\bsnm{Lee}, \binits{J.}}:
Understanding catastrophic overfitting in single-step adversarial training.
arXiv preprint arXiv:2010.01799
(2020)
\end{botherref}
\endbibitem

\bibitem{MMA}
\begin{botherref}
\oauthor{\bsnm{Ding}, \binits{G.W.}},
\oauthor{\bsnm{Sharma}, \binits{Y.}},
\oauthor{\bsnm{Lui}, \binits{K.Y.C.}},
\oauthor{\bsnm{Huang}, \binits{R.}}:
Mma training: Direct input space margin maximization through adversarial
  training.
arXiv preprint arXiv:1812.02637
(2018)
\end{botherref}
\endbibitem

\bibitem{IAAT}
\begin{botherref}
\oauthor{\bsnm{Balaji}, \binits{Y.}},
\oauthor{\bsnm{Goldstein}, \binits{T.}},
\oauthor{\bsnm{Hoffman}, \binits{J.}}:
Instance adaptive adversarial training: Improved accuracy tradeoffs in neural
  nets.
arXiv preprint arXiv:1910.08051
(2019)
\end{botherref}
\endbibitem

\bibitem{MDP}
\begin{barticle}
\bauthor{\bsnm{Puterman}, \binits{M.L.}}:
\batitle{Markov decision processes}.
\bjtitle{Handbooks in operations research and management science}
\bvolume{2},
\bfpage{331}--\blpage{434}
(\byear{1990})
\end{barticle}
\endbibitem

\bibitem{DQN}
\begin{botherref}
\oauthor{\bsnm{Mnih}, \binits{V.}},
\oauthor{\bsnm{Kavukcuoglu}, \binits{K.}},
\oauthor{\bsnm{Silver}, \binits{D.}},
\oauthor{\bsnm{Graves}, \binits{A.}},
\oauthor{\bsnm{Antonoglou}, \binits{I.}},
\oauthor{\bsnm{Wierstra}, \binits{D.}},
\oauthor{\bsnm{Riedmiller}, \binits{M.}}:
Playing atari with deep reinforcement learning.
arXiv preprint arXiv:1312.5602
(2013)
\end{botherref}
\endbibitem

\bibitem{DDPG}
\begin{botherref}
\oauthor{\bsnm{Lillicrap}, \binits{T.P.}},
\oauthor{\bsnm{Hunt}, \binits{J.J.}},
\oauthor{\bsnm{Pritzel}, \binits{A.}},
\oauthor{\bsnm{Heess}, \binits{N.}},
\oauthor{\bsnm{Erez}, \binits{T.}},
\oauthor{\bsnm{Tassa}, \binits{Y.}},
\oauthor{\bsnm{Silver}, \binits{D.}},
\oauthor{\bsnm{Wierstra}, \binits{D.}}:
Continuous control with deep reinforcement learning.
arXiv preprint arXiv:1509.02971
(2015)
\end{botherref}
\endbibitem

\bibitem{TD3}
\begin{bchapter}
\bauthor{\bsnm{Fujimoto}, \binits{S.}},
\bauthor{\bsnm{Hoof}, \binits{H.}},
\bauthor{\bsnm{Meger}, \binits{D.}}:
\bctitle{Addressing function approximation error in actor-critic methods}.
In: \bbtitle{International Conference on Machine Learning},
pp. \bfpage{1587}--\blpage{1596}
(\byear{2018}).
\bcomment{PMLR}
\end{bchapter}
\endbibitem

\bibitem{TRPO}
\begin{bchapter}
\bauthor{\bsnm{Schulman}, \binits{J.}},
\bauthor{\bsnm{Levine}, \binits{S.}},
\bauthor{\bsnm{Abbeel}, \binits{P.}},
\bauthor{\bsnm{Jordan}, \binits{M.}},
\bauthor{\bsnm{Moritz}, \binits{P.}}:
\bctitle{Trust region policy optimization}.
In: \bbtitle{International Conference on Machine Learning},
pp. \bfpage{1889}--\blpage{1897}
(\byear{2015}).
\bcomment{PMLR}
\end{bchapter}
\endbibitem

\bibitem{PPO}
\begin{botherref}
\oauthor{\bsnm{Schulman}, \binits{J.}},
\oauthor{\bsnm{Wolski}, \binits{F.}},
\oauthor{\bsnm{Dhariwal}, \binits{P.}},
\oauthor{\bsnm{Radford}, \binits{A.}},
\oauthor{\bsnm{Klimov}, \binits{O.}}:
Proximal policy optimization algorithms.
arXiv preprint arXiv:1707.06347
(2017)
\end{botherref}
\endbibitem

\bibitem{SAC}
\begin{bchapter}
\bauthor{\bsnm{Haarnoja}, \binits{T.}},
\bauthor{\bsnm{Zhou}, \binits{A.}},
\bauthor{\bsnm{Abbeel}, \binits{P.}},
\bauthor{\bsnm{Levine}, \binits{S.}}:
\bctitle{Soft actor-critic: Off-policy maximum entropy deep reinforcement
  learning with a stochastic actor}.
In: \bbtitle{International Conference on Machine Learning},
pp. \bfpage{1861}--\blpage{1870}
(\byear{2018}).
\bcomment{PMLR}
\end{bchapter}
\endbibitem

\bibitem{MNIST}
\begin{botherref}
\oauthor{\bsnm{LeCun}, \binits{Y.}}:
The mnist database of handwritten digits.
http://yann. lecun. com/exdb/mnist/
(1998)
\end{botherref}
\endbibitem

\bibitem{CIFAR-10}
\begin{botherref}
\oauthor{\bsnm{Krizhevsky}, \binits{A.}},
\oauthor{\bsnm{Hinton}, \binits{G.}}, et al.:
Learning multiple layers of features from tiny images
(2009)
\end{botherref}
\endbibitem

\bibitem{SVHN}
\begin{botherref}
\oauthor{\bsnm{Netzer}, \binits{Y.}},
\oauthor{\bsnm{Wang}, \binits{T.}},
\oauthor{\bsnm{Coates}, \binits{A.}},
\oauthor{\bsnm{Bissacco}, \binits{A.}},
\oauthor{\bsnm{Wu}, \binits{B.}},
\oauthor{\bsnm{Ng}, \binits{A.Y.}}:
Reading digits in natural images with unsupervised feature learning
(2011)
\end{botherref}
\endbibitem

\bibitem{Pre-Act_Resnet-18}
\begin{bchapter}
\bauthor{\bsnm{He}, \binits{K.}},
\bauthor{\bsnm{Zhang}, \binits{X.}},
\bauthor{\bsnm{Ren}, \binits{S.}},
\bauthor{\bsnm{Sun}, \binits{J.}}:
\bctitle{Identity mappings in deep residual networks}.
In: \bbtitle{European Conference on Computer Vision},
pp. \bfpage{630}--\blpage{645}
(\byear{2016}).
\bcomment{Springer}
\end{bchapter}
\endbibitem

\bibitem{CLR}
\begin{bchapter}
\bauthor{\bsnm{Smith}, \binits{L.N.}}:
\bctitle{Cyclical learning rates for training neural networks}.
In: \bbtitle{2017 IEEE Winter Conference on Applications of Computer Vision
  (WACV)},
pp. \bfpage{464}--\blpage{472}
(\byear{2017}).
\bcomment{IEEE}
\end{bchapter}
\endbibitem

\end{thebibliography}

\end{document}